# Transfer Learning Approach for Arabic Offensive Language Detection System

BERT-Based Model

Fatemah Husain
Department of Information Science, Kuwait University
Kuwait City, State of Kuwait
f.husain@ku.edu.kw

Ozlem Uzuner
Department of Information Sciences and Technology
George Mason University
Fairfax, USA
ouzuner@gmu.edu

*Abstract*—The problem of online offensive language limits health and security of online users. It is very important to apply the latest state-of-the-art techniques in developing a system to detect online offensive language and to ensure social justice to the online communities. Our study investigates the effects of transfer learning across several Arabic offensive language datasets. We develop multiple classifiers that use four datasets individually and combined to gain knowledge about online Arabic offensive content and classify users comments accordingly. Our results demonstrate the limited effects of transfer learning on the performance of the classifiers, particularly for highly dialectic comments.

## I. INTRODUCTION

Developing a system to detect online offensive language is very important to the health and the security of online users. Studies have shown that cyberhate, online harassment and other misuses of technology are on the rise. Particularly during the global Coronavirus pandemic in 2020, 35% reported online harassment related to their identity-based characteristics, which is a 3% increase over 2019[1].

Applying advanced techniques from the Natural Language Processing (NLP) field to support the development of an online hate free community is a critical task for social justice. Transfer learning enhances the performance of the classifier by allowing the transfer of knowledge from one domain or one dataset to others that have not been seen before, thus, supporting the classifier to be more generalizable. In our study, we apply the principles of transfer learning cross multiple Arabic offensive language datasets to compare the effects on system performance.

This study aims at investigating the effects of fine-tuning and training Bidirectional Encoder Representations from Transformers (BERT) model on multiple Arabic offensive language datasets individually and testing it using other datasets individually. Our experiment starts with a comparison among multiple BERT models to guide the selection of the main model for our study. The study also investigates the effects of concatenating all datasets to be used for fine-tuning and training BERT model.

## II. BACKGROWND

### A. Arabic Language

The scope of this study covers Arabic text from online user-generated content. While there are multiple forms of the Arabic language, the majority of the content from user-generated platforms are written in dialectic Arabic. The dialectic form of Arabic is the actual spoken Arabic, and it has several categories depending on social and geographical factors. Habash [1] divides the Arabic dialects into seven categories; Egyptian, Levantine, Gulf, North African, Iraqi, Yemenite, and Maltese. The diversity among Arabic dialects adds difficulties to the process of developing an NLP system that can understand online Arabic content similar to human level of understanding.

### B. Offensive Language

Offensive language could include abusive behavior, actions with the intention of harming others, threats, discrimination against people, swear words, blunt insults, hate speech, aggressive content, cyberbullying, and toxic comments [2; 3]. The exchange of offensive language online creates an environment of disturbance, disrespect, anger, affecting the harmony of conversations, and reduces the trust of users to the online platform [4].

### C. Related Work

BERT stands for Bidirectional Encoder Representations from Transformers. It is an innovative language model that presents state-of-the-art results in multiple NLP tasks, such as question answering and language inference. BERT applies pre-trained language representations to down-stream tasks through a fine-tuning approach to learn the language representations. This approach is also called transfer learning, in which the pre-trained language representations is developed using a neural network model on a known task, and then a fine-tuning technique is performed to use the same model for a new purpose-specific task. The main feature that distinguish BERT from the other language modeling techniques is the use of bidirectional language model rather than unidirectional language model during the fine-tuning process. This bidirectional learning technique consists of a Masked Language Model (MLM) with a pre-training objective that is randomly masks some of the tokens from the input with the objective of predicting the original vocabulary id of the masked word based only on its context [5].

---
[1] https://www.adl.org/

Multilingual BERT (M-BERT) [2] proposed by Google Research and has 2 versions; BERT-base-multilingual-uncased model, which covers 102 languages, and BERT-base-multilingual-cased model, which covers 104 languages. Wikipedia dumps of each language (excluding user and talk pages) were used to train the models with a shared word piece vocabulary. Previous studies report that M-BERT outperforms other tools when it applies to multilingual text, however, M-BERT shows some limitations in tokenizing Arabic sentences, which could degrade the performance of the classifier [6]. This finding is in line with other experiments conducted by Hasan et al. [7], Saeed et al. [8], and Keleg et al. [9], which reported poor performance in Arabic offensive language and hate speech detection in comparison to other word embeddings, machine learning classifiers, and deep learning classifiers. In addition, Abu Farha and Magdy [10] tried M-BERT with Adam optimiser, and trained the model with 4 epochs, learning rate of 1e−5, and setting the maximum sequence length to the maximum length seen in the training set. However, the results were not as good as the results obtained from the Bidirectional Long Short-Term Memory (BiLSTM) model, Convolutional Neural Network - Bidirectional Long Short-Term Memory (CNN-BiLSTM) model, and multitask learning models. In [11], multiple M-BERT-based classifiers were used with different fine-tuning settings for offensive language and hate speech detection tasks, and in both tasks the reported macro F1 score was not better than what has been reported by other studies using simple traditional machine learning methods [12].

While M-BERT supports various languages, Arabic specific BERT models have been used as well for Arabic offensive language detection, such as AraBERT and PERT-base Arabic. The AraPERT[3] is an Arabic version of BERT model that shows state-of-the-art performance in multiple downstream tasks [13]. It uses BERT-base configuration has similar pre-training settings for the ones used at the original BERT model, consisting of implementing the Masked Language Modeling (MLM) task and the Next Sentence Prediction (NSP) task. Multiple Modern Standard Arabic (MSA) corpora are used to train the model, which include: manually scraped Arabic news websites for articles, 1.5 billion words extracted from news articles from ten major news sources, and OSIAN, which is an Open Source International Arabic News Corpus. Results of evaluating AraBERT on sentiment analysis task, question answering task, and Named Entity Recognition (NER) task outperform others for all tasks from M-BERT and from the previous state-of-the-art models archived by Dahou et al. [14] and Eljundi et al. [6]. This finding demonstrates that a pre-trained language model trained on a single language performs better than a multilingual model.

Djandji et al. [15] apply AraBERT for the Open-Source Arabic Corpora and Corpora Processing Tools (OSACT) dataset with a multitask learning approach and a multilabel classification approach. Multitask Learning solves the data imbalance problem in OSACT dataset by leveraging information from multiple tasks simultaneously. The same study also applies multilabel classification approach using AraBERT, in which all labels of the 2 labeling hierarchies in OSACT dataset—offensive and hate—are merged under a broad task of violence detection. Results report 90.15% as the highest macro F1 score for offensive language detection when adopting the multitask learning approach with AraBERT. Findings from this study demonstrate the superiority of using a multitask learning approach over a multilabel classification approach when using AraBERT for offensive language detection. The error analysis reveals that confusion occurs in tweets that consist of offensive words in a non-offensive context. It also shows that most of the errors are related to mockery, sarcasm, or mentioning other offensive and hateful statements within tweets.

Arabic-base-BERT[4] model is another Arabic monolingual BERT model [16]. The pre-trained corpus consists of multiple Arabic resources such as Arabic OSCAR and Arabic Wikipedia, which includes MSA and dialectic Arabic. Results of evaluating the performance of the model for sentiment analysis task shows higher F1 score for Arabic-base-BERT model than M-BERT model and hULMonA[5] model (state of the art model for Arabic sentiment analysis) when used with Levantine dialect and Egyptian dialect datasets.

III. METHODOLOGY

A. Datasets

We use four publicly available Arabic Offensive language datasets. These datasets include: Aljazeera.net Deleted Comments [17], YouTube dataset [18], Levantine Twitter Dataset for Hate Speech and Abusive Language (L-HSAB) [19], and OSACT offensive and not offensive classification samples [20]. Table 1 provides a summary for the characteristics of each dataset.

We use only binary classes; offensive or not offensive. Thus, we convert different types of offensive languages to offensive class. For example, the L-HSAB dataset differentiate between hate and abusive languages classes; which were both converted to offensive language class.

---

[2] https://github.com/google-research/bert/blob/master/multilingual.md
[3] https://github.com/aub-mind/arabert
[4] https://github.com/alisafaya
[5] https://github.com/aub-mind/hULMonA

For some datasets that are provided in a train/evaluation/test splitted formats, we merge all parts together into one dataset, and then, we randomly apply 80%-20% split for train-test datasets. This support consistency in the setting among all datasets used in this study as most of them are provided in one part. All datasets were used without performing any preprocessing procedures to the text.

TABLE I. DATASET DESCRIBTION

| Dataset | Source | Labels/Size |
|---|---|---|
| Aljazeera Deleted Comments (Mubarak et al., 2017) | Aljazeera News (Aljazeera.net) | 31,692 comments (offensive = 25,506, clean = 5,653, obscene = 533) |
| YouTube Comments (Alakrot, Murray, & Nikolov, 2018) | YouTube | 15,050 comments (not offensive = 9,237, offensive = 5,813) |
| Levantine Twitter Dataset for Hate Speech and Abusive Language (L-HSAB) (Mulki et al., 2019) | Twitter | 5,846 tweets (hate = 468, abusive = 1,728, normal = 3,650) |
| Open-Source Arabic Corpora and Corpora Processing Tools (OSACT) (Mubarak et al., 2020) | Twitter | 10,000 tweets (offensive = 1,900, not offensive = 8,100) (hate = 500, not hate = 9,500) |

B. *BERT Models*

Our experiments depend mainly on AraBERT model from Hugging Face[6] library. To ensure selecting the best available BERT model for our task, we use the OSACT and the L-HSAB datasets to perform a quick experiment and evaluate the performance of multiple BERT models that are supporting Arabic to guide our selection for the model. We use XLM-Roberta[7] (also called XLM-R), M-BERT, Arabic-Base-BERT, and AraBERT. Table 2 shows the resulted macro F1 scores of the experiments. As can be noticed from table 2, AraBERT outperforms the other models. Thus, we decide to select AraBERT for our main experiments. Moreover, table also demonstrates that Arabic monolingual models perform better than multilingual models.

In all experiments, we apply the same experiment settings: maximum length = 128, patch size = 16, epoch = 5, epsilon = 1e-8, and learning rate = 2e-5. We use the pooled output from the encoder to be used with a simple Feed Forward Neural Network (FFNN) layer to build the classifier. Experiments were developed in Python using PyTorch-Transformers library, and evaluation metrics were developed using Scikit-Learn Python library. Google Colab Pro used to conduct all experiments.

---
[6] https://huggingface.co/
[7] https://github.com/pytorch/fairseq/tree/master/examples/xlmr

TABLE II. F1 SCORES FOR EXPERIMENTAL STUDIES OF DIFFERENT BERT MODELS

| Dataset | Arabic Monolingual Model | | Multilingual Model | |
|---|---|---|---|---|
| | AraBERT | Arabic-base-BERT | Multilingual BERT | XLM-RoBERTa |
| OSACT-Offensive | **0.90** | 0.88 | 0.87 | 0.86 |
| L-HSAB | **0.72** | 0.69 | 0.61 | 0.44 |

IV. RESULTS AND DISCUSSION

A. *Results*

We use macro measurements of precision, recall, and F1, in addition to accuracy score to evaluate the performance of the classifiers. The following table shows performance results for the four individual models, each fine-tuned and trained using one dataset and tested on all datasets individually.

TABLE III. PERFORMANCE RESULTS

| Train Dataset | Test Dataset | Precision | Recall | F1 | Accuracy |
|---|---|---|---|---|---|
| OSACT | OSACT | **0.91** | **0.91** | **0.91** | **0.94** |
| | L-HSAB | 0.77 | 0.77 | 0.77 | 0.78 |
| | YouTube | 0.74 | 0.78 | 0.75 | 0.78 |
| | Aljazeera | 0.69 | 0.62 | 0.52 | 0.54 |
| L-HSAB | OSACT | 0.83 | 0.74 | 0.77 | 0.82 |
| | L-HSAB | **0.87** | **0.87** | **0.87** | **0.88** |
| | YouTube | 0.77 | 0.78 | 0.77 | 0.78 |
| | Aljazeera | 0.61 | 0.59 | 0.39 | 0.39 |
| YouTube | OSACT | 0.86 | 0.8 | 0.83 | **0.88** |
| | L-HSAB | 0.82 | 0.82 | 0.82 | 0.83 |
| | YouTube | **0.87** | **0.87** | **0.87** | **0.88** |
| | Aljazeera | 0.68 | 0.61 | 0.51 | 0.53 |
| Aljazeera | OSACT | 0.76 | 0.67 | 0.6 | 0.63 |
| | L-HSAB | 0.64 | 0.7 | 0.54 | 0.56 |
| | YouTube | **0.92** | 0.71 | 0.65 | 0.65 |
| | Aljazeera | 0.73 | **0.75** | **0.74** | **0.85** |

As can be noticed from the table above, the highest recorded macro recall, macro F1, and accuracy scores are shown for the OSACT dataset when used in training and testing. While the highest recorded macro precision is reported by the classifier that has been trained using the Aljazeera dataset and tested with YouTube dataset. It is also noticeable that Aljazeera dataset is the one with the lowest overall performance scores. Results from each individual dataset experiments demonstrate highest performance when the model is trained and tested using the same dataset, which indicates the limited improvements of the transfer learning approach.

Table 4 shows the results after concatenating all datasets into one corpus, and then use it to fine-tune and train the classifier. Comparing the best results for each dataset from table 3 with the results from the concatenating model, overall there are no improvement achieved. The OSACT dataset is still recording the highest performance scores, which are exactly the same as if the model is fine-tuned and trained using the OSACT dataset only. However, the L-HSAB dataset shows lower performance by 3% in macro F1 score. This decrease could be a result of the high dialectic text of L-HSAB, as the other datasets might not share much Levantine vocabulary words.

TABLE IV. PERFORMANCE RESULTS FOR THE CONCATENATED TRAINED MODEL

| Test Dataset | Precision | Recall | F1 | Accuracy |
|---|---|---|---|---|
| OSACT | **0.9** | **0.91** | **0.9** | **0.94** |
| L-HSAB | 0.86 | 0.84 | 0.84 | 0.85 |
| YouTube | 0.87 | 0.87 | 0.87 | 0.88 |
| Aljazeera | 0.74 | 0.75 | 0.74 | 0.85 |

### B. Error Analysis

Looking over samples of the misclassified comments, we calculate percentages of misclassified comments per class for each experiment that was conducted using the concatenated trained model. Table 5 presents a summary of the error analysis. The percentages are calculated per class based on the total instances of each class for each of the four testing datasets. Most common 5 tokens are presented in the table based on their frequencies order.

As can be seen from table 5, offensive and not offensive misclassified percentages vary among the datasets. Investigating top tokens among the misclassified samples shows names of countries (e.g. Saudi Arabia, Qatar, Iraq) and names of famous people (e.g. Kadim, Ahlam, Gibran), which are in some cases refer to the first name and the last name of the same person as two separate tokens. For example, 'Kadim' and 'Sahir' are the first and the last name of the same singer, and 'Gibran' and 'Basil' are the first name and the last name of the same minster. This type of terms need to be proceeded as one term rather than separate parts because the semantic meaning of its parts might not be equivalent to the semantic meaning of the term. As a result of that, some preprocessing procedures are needed to ensure proper understanding and processing of multiple tokens terms.

TABLE V. ERROR ANALYSIS SUMMARY FOR THE CONCATENATED TRAINED MODEL

| | Misclassified % | | Misclassified Top Common Tokens | |
|---|---|---|---|---|
| Test Dataset | Offensive | Not Offensive | Offensive | Not Offensive |
| OSACT | 16% | 3.5% | الله/God<br>علوقيه/stereotype<br>صغير/small<br>بنت/girl<br>حبيبي/lovely | الله/God<br>😂<br>عيون/eyes<br>ناس/people<br>موضوع/topic |
| L-HSAB | 8.7% | 19.4% | جبران/Gibran<br>باسيل/Basil<br>حلوة/beautiful<br>وزير/minister<br>نعيم/blessing | جبران/Gibran<br>باسيل/Basil<br>سوريا/Syria<br>قطر/Qatar<br>دولة/state |
| YouTube | 15% | 11% | كاظم/Kadim<br>الله/God<br>احلام/Ahlam<br>ساهر/Sahir<br>عراقي/Iraqi | كاظم/Kadim<br>الله/God<br>ناس/people<br>مصر/Egypt<br>ساهر/Sahir |
| Aljazeera | 8.2% | 44.7% | الله/God<br>دولة/state<br>عراق/Iraq<br>شعب/nation<br>جزيرة/Jazeera | الله/God<br>جزيرة/Jazeera<br>سعودية/Saudi Arabia<br>دولة/state<br>مسلمين/muslims |

Since our system is depending on the vocabulary list of AraBERT model, further investigation to AraBERT training corpus is important to consider. AraBERT is trained mostly on Arabic News outlet, the followings are the sources of its training raw text:

1. Arabic Wikipedia database dump[8]
2. 1.5B words Arabic Corpus [9] (sources include newspapers, books, and research papers)
3. Open Source International Arabic News Corpus (OSIAN)[10]
4. Assafir Lebanese news articles[11]
5. Manually crawled news websites (Al-Akhbar, Annahar, AL-Ahram, and AL-Wafd).

The sources of our datasets are mostly from user-generated content, which differ from the Arabic text that is used in writing news articles and books. The type of Arabic that is used in our datasets is the dialectic Arabic, while the one used in training AraBERT is the MSA. Thus, simple fine-tuning process might not be enough to adjust the weights of AraBERT vocabulary toward our task of offensive language detection, especially if most of the tokens in our datasets are treated as out-of-vocabulary tokens by AraBERT tokenizer.

### C. System Implications

Increasing the dataset size for training and fine-tuning AraBERT model not always improve the performance of the system. Thus, finding some other methods to improve the performance are required. For example, creating more advance classifier architecture on top of BERT model that can give better results than a simple FFNN. Another method could focus on AraBERT model and trying to adjust its vocabulary to support offensive language classification task. A costlier approach could be to consider traning a new BERT model that is customized for the online Arabic offensive detection task.

## V. CONCLUSION

In this paper, we try to present our work in applying transfer learning cross several Arabic offensive detection datasets using AraBERT model. Our results report outperformance of Arabic monolingual BERT models over BERT multilingual models. The results also report poor performance when applying transfer learning cross individual datasets from heterogeneous sources and themes; such as YouTube comments from musicians' channels and Aljazeera News comments from political articles. While the results from aggregating knowledge from multiple datasets on the same time show no effects on the performance when tested on individual datasets, it lowers the performance of the highly dialectic dataset; L-HSAB; by 3% in macro F1 score. The overall findings from our experiments demonstrate the importance of developing novel methods to pre-train BERT model for the tasks of Arabic offensive language detection.

---

[8] https://archive.org/details/arwiki-20190201
[9] https://arxiv.org/pdf/1611.04033.pdf
[10] https://www.aclweb.org/anthology/W19-4619/
[11] http://m.assafir.com/Channel/50/English/TopMenu